%% file: main.tex
\documentclass[10pt,twocolumn,letterpaper]{article}

 \usepackage{cvpr}              


\usepackage{amsmath}
\usepackage{amssymb}
\usepackage{amsthm}
\usepackage{amsfonts}
\usepackage{booktabs}
\usepackage{tabularx}
\newtheorem*{remark}{Remark}
\usepackage{pifont}
\usepackage{enumitem}
\setlist[description]{leftmargin=1em} 
\usepackage{colortbl}
\def\redc{\cellcolor[HTML]{FF999A}}
\def\orangec{\cellcolor[HTML]{FFCC99}}
\def\yellowc{\cellcolor[HTML]{FFF8AD}}

\definecolor{cvprblue}{rgb}{0.21,0.49,0.74}
\usepackage[pagebackref,breaklinks,colorlinks,allcolors=cvprblue]{hyperref}

\def\paperID{11705} 
\def\confName{CVPR}
\def\confYear{2025}

\title{NSA: Neuro-symbolic ARC Challenge}

\author{
    Paweł Batorski\textsuperscript{1} \quad
    Jannik Brinkmann\textsuperscript{2} \quad
    Paul Swoboda\textsuperscript{1} \\
    {\tt\small pawel.batorski@hhu.de \quad jannik.brinkmann@uni-mannheim.de \quad paul.swoboda@hhu.de} \\
    \textsuperscript{1}Heinrich Heine Universität Düsseldorf \quad
    \textsuperscript{2}University of Mannheim
}

\begin{document}
\twocolumn[{%
\renewcommand\twocolumn[1][]{#1}%
\maketitle
\begin{center}
    \centering
    \captionsetup{type=figure}
    \includegraphics[width=\textwidth, keepaspectratio]{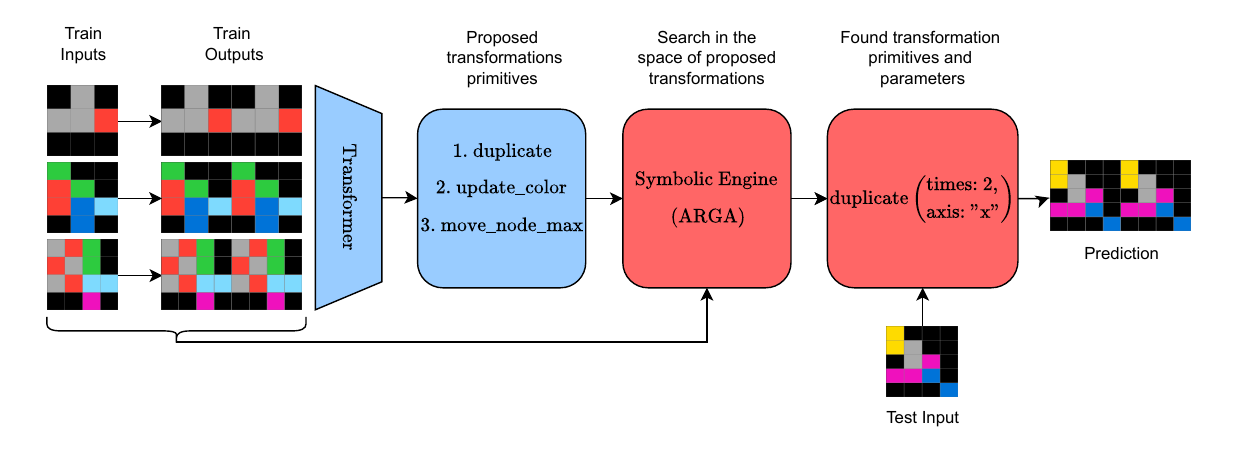}
    \captionof{figure}{An overview of the NSA framework: Starting with input-output pairs, a transformer model is utilized to propose potential transformation primitives. These are then feed to the symbolic combinatorial search (ARGA~\cite{xu2023graphs}), which identifies the correct overall transformation and corresponding parameters solving the task for the given training input-output pairs.
    Finally, the selected transformation is applied to the test input to generate the final prediction.}
    \label{fig:teaser}
\end{center}%
}]

\input{sec/abstract_arxiv}    
\input{sec/intro}
\input{sec/related_work}
\input{sec/method}
\input{sec/experiments}
\input{conclusion}


\end{document}

%% file: sec/abstract_arxiv.tex
\begin{abstract}
The Abstraction and Reasoning Corpus (ARC) evaluates general reasoning capabilities that are difficult for both machine learning models and combinatorial search methods. 
We propose a neuro-symbolic approach that combines a transformer for proposal generation with combinatorial search using a domain-specific language. 
The transformer narrows the search space by proposing promising search directions, which allows the combinatorial search to find the actual solution in short time.
We pre-train the trainsformer with synthetically generated data.
During test-time we generate additional task-specific training tasks and fine-tune our model. 
Our results surpass comparable state of the art on the ARC evaluation set by 27\% and compare favourably on the ARC train set.
We make our code and dataset publicly available at \href{https://github.com/Batorskq/NSA}{https://github.com/Batorskq/NSA}.
\end{abstract}

%% file: sec/intro.tex
\section{Introduction}
\label{sec:intro}

\begin{figure*}[ht]
    \centering
    \includegraphics[width=0.98\textwidth]{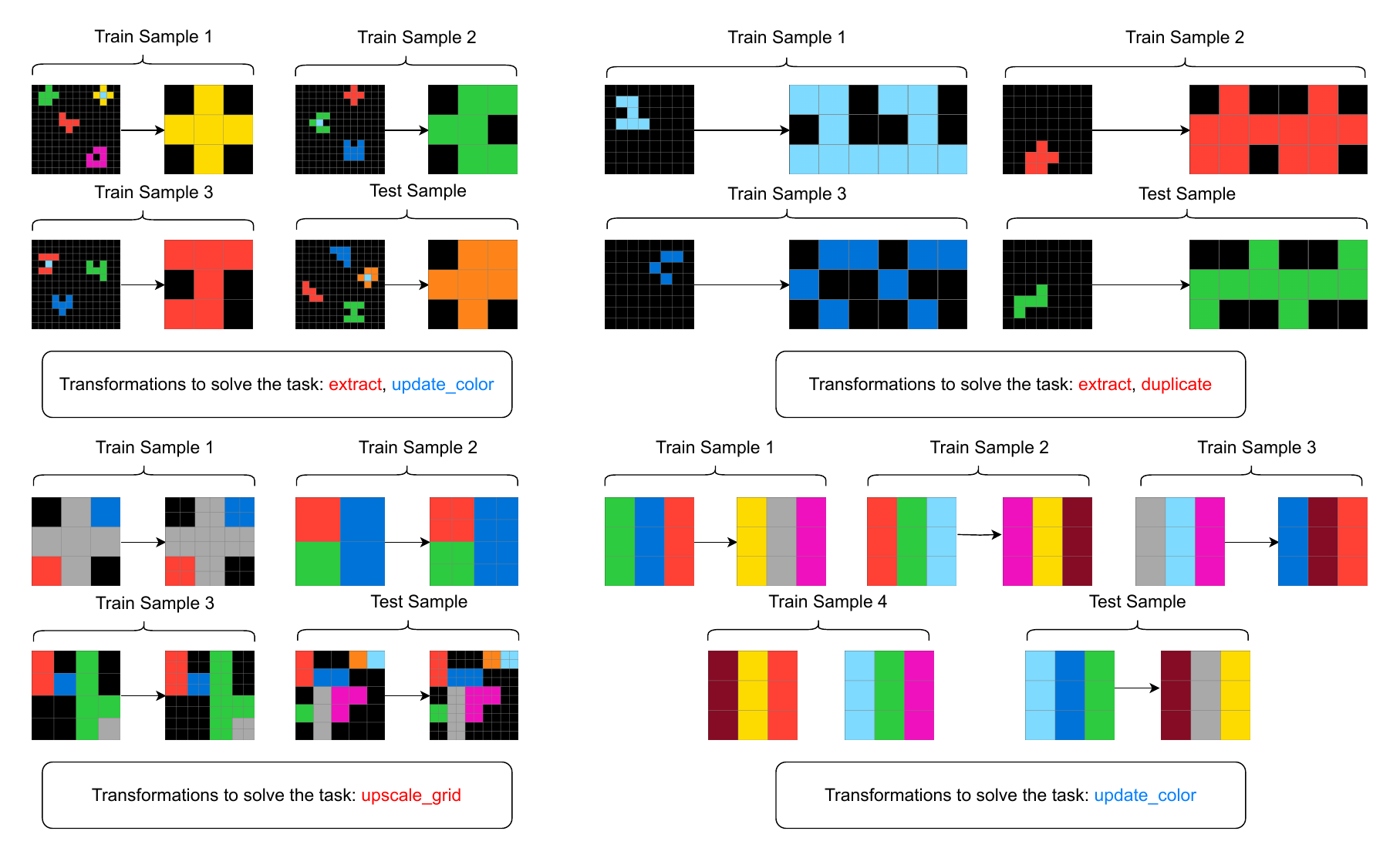}
    \vspace{1em} 

    \begin{subtable}[c]{0.48\textwidth}
        \centering
        \small 
        \begin{tabular}{@{}ccc@{}}
            \toprule
            & \begin{tabular}{@{}c@{}}DSL Rep. \\ Capacity\end{tabular} 
            & \begin{tabular}{@{}c@{}}Effective \\ Capacity\end{tabular}  \\
            \midrule
            ARGA & \ding{55} & \ding{55} \\
            ARGAe & \ding{51} & \ding{55} \\
            NSA & \ding{51} & \ding{51} \\
            \bottomrule
        \end{tabular}
        \label{tab:model_capacities}
    \end{subtable}

    \caption{
Input-output pairs and a test image from selected ARC tasks, each with a distinct transformation challenge.
Our solution is illustrated for each task.
ARGA cannot solve any of the first three tasks since it cannot represent the needed transformation.
Our extended DSL ARGAe has the representational capacity but can run into time-out during search and thus cannot find it effectively in practice.
For the bottom right task ARGA can represent the transformation theoretically but neither ARGA nor ARGAe can find it due to a time-out during search.
Necessary transformation primitives absent in ARGA’s DSL are marked in red: \textcolor{red}{extract}, \textcolor{red}{duplicate}, and \textcolor{red}{upscale\_grid}.
In contrast, transformations present in ARGA's original DSL, like \textcolor{blue}{update\_color}, are highlighted in blue.
For the tasks in the upper right and lower left corner all necessary transformation primitives are missing in ARGA, while for the upper left one transformation primitive is present and another is lacking. 
NSA addresses this dilemma by proposing the necessary transformation primitives to explore during combinatorial search.
}
\label{fig:showcase_and_table}
\end{figure*}

The Abstraction and Reasoning Corpus (ARC) challenge~\cite{chollet2019measureintelligence} is a difficult few-shot benchmark for testing visual reasoning capabilities of machine learning models.
The capabilities of recent general-purpose LLM systems are, as of now, not good enough to solve ARC at human performance in a reasonably limited amount of time~\cite{xu2023llms,mirchandani2023largelanguagemodelsgeneral,mitchell2023comparing}.
Arguably their pre-training seems to have not imbued them with enough of the necessary concepts required to solve ARC tasks reliably and without an excessive number of tries.
It is unclear whether LLMs lack the correct level of abstraction and the specific type of high-level visual reasoning used for solving visual riddles.
On the other hand, specialized combinatorial search methods using domain specific languages (DSL)~\cite{ferre2024tackling,ainooson2023approach,xu2023graphs,lei2024generalized} struggle with the great variety and generality of the ARC tasks, with no compact enough description of the search space being available to effectively solve the problem via traditional methods.
At the same time, tasks in the ARC benchmark are not hard for humans, with most tasks seemingly requiring only some visual pattern matching and a limited amount of reasoning. In user studies, human participants solved around 80\% of the tasks on average~\cite{johnson2021fast}.

This indicates that (i)~currently pure pre-trained ML systems are not yet capable of solving ARC with reasonably limited compute and may lack some fundamental capabilities to do so and (ii)~specialized combinatorial approaches relying on DSLs cannot cope with the search space complexity.
However, both approaches can be used to capture some aspects of humans solving ARC tasks well:
DSLs can be used for reasoning on the right abstraction level in terms of composing elementary transformations.
ML can be used to pick promising transformation candidates, mimicking the intuitive way a human will restrict his reasoning instead of enumerating all possible transformations.
We propose a neuro-symbolic approach to solving ARC that combines these strengths of ML and DSL-based combinatorial methods.
In our approach the DSLs give us valuable inductive bias that allows us to reason on the right abstraction level, while the ML part guides the search process by proposal generation, allowing the combinatorial search to actually find solutions in a limited amount of time.

To train our neuro-symbolic approach, the provided ARC training data is not enough. 
The training set consists of only 400 tasks, mostly with three or four input/output pairs each.
Due to this limited training data, we synthetically generate more training data to pre-train and test-time fine-tune our proposal generation transformer.

Our empirical results indicate that our combined approach can surpass pure ML- and combinatorial methods.

\paragraph{Contributions:}
In detail, our contributions are
\begin{description}
    \item[DSL:] We extend the DSL from~\cite{xu2023graphs} to increase its representational capacity, enabling it to solve significantly more ARC tasks.
    \item[Transformer-Guided Neuro-Symbolic Search:] We train a transformer model to suggest suitable transformation primitives that restrict the search process.
    We pre-train our model on a large set of ARC tasks synthetically generated by hind-sight relabeling.
    During inference time, we fine-tune our model by again synthetically generating ARC tasks using only the currently considered task images.
    The transformer then generates proposals that are fed to the combinatorial DSL search.
    \item[Experiments:] We demonstrate the efficacy of our approach on the ARC train and evaluation set. On the evaluation set we outperform comparable baselines, i.e.\ the pure DSL methods~\cite{xu2023graphs,ferre2024tackling,ainooson2023approach} model as well as recent ML approaches~\cite{butt2024codeit,mirchandani2023largelanguagemodelsgeneral}.
\end{description}

%% file: sec/related_work.tex
\section{Related Work}

The ARC challenge~\cite{chollet2019measureintelligence} was proposed to benchmark general reasoning capabilities that seem easy for humans but hard for current ML methods and hand-crafted search algorithms.
It has lead to a large number of attempts using various paradigms for its solution.
To spur progress, leaderboards have been made available that measure performance of submitted systems.
For inclusion in the primary leaderboard, the model may not call any outside or proprietary LLM and a 30-minute computation time limit for each task must be obeyed.
For models that do not adhere to these requirements, a secondary leaderboard can be used.

We categorize related work on solving the ARC challenge into learning based and DSL-based approaches and also discuss simpler alternatives to full ARC.

\subsection{ML}
The work \cite{xu2023llms} analyzes one-shot prompting of large language models (LLMs) for ARC Challenge problems. The authors examine the performance of LLaMA 2 \cite{touvron2023llama} and GPT-4 \cite{achiam2023gpt} with various grid representations to address ARC tasks. By applying few-shot and in-context few-shot prompting, they solve only 13 tasks from a chosen subset of ARC tasks that can be easily addressed by the DSL approach in~\cite{xu2023graphs}. However, incorporating the abstract graph reasoning proposed in~\cite{xu2023graphs} improves their results, enabling them to solve 23 out of 50 tasks. While this graph abstraction significantly enhances the LLMs' reasoning capabilities on ARC tasks, the overall performance remains far below the results achieved with simple abstraction-based reasoning alone.
The work~\cite{mitchell2023comparing} explores various methods of representing ARC images in LLMs.

The work~\cite{kolev2020neuralabstractreasoner} uses a differential neural computer to solve ARC problems with a grid size of up to $10 \times 10$.
The approach~\cite{alford2022neural} uses DreamCoder~\cite{ellis2023dreamcoder} for solving a small subset of ARC tasks.
An imitation learning algorithm based on the decision transformer~\cite{chen2021decision} and a custom clustering method is used in~\cite{park2023unravelingarcpuzzlemimicking} for solving small ARC tasks.
Due to scaling or representational capacity issues the above works are all applied to small subsets of ARC and do not aim to solve the full ARC dataset.

The works~\cite{mirchandani2023largelanguagemodelsgeneral} and~ \cite{wang2024hypothesis} use pre-trained LLMs for directly outputting the transformation of the test input image when conditioned on each task's input output pairs.
Greenblatt's pure LLM method~\cite{greenblatt2024getting} used GPT-4o~\cite{openai2024gpt4ocard} and few-shot prompting to sample python programs for describing ARC transformations. For each task up to 8000 programs are sampled. While achieving impressive results, this puts the approach outside the evaluation protocol of using 30 minutes per task and not using a closed source LLM.
The vision transformer~\cite{dosovitskiy2021an}-based work~\cite{li2024tackling} uses the spatial image structure and trains on procedurraly generated tasks~\cite{hodel2024rearc}. However, an evaluation on the ARC train or eval set is missing.

The closest work to ours is~\cite{butt2024codeit}, which uses a large language model fine-tuned on code generation to write programs using the DSL~\cite{hodel2023domain}.
Similar to us, the transformer is fine-tuned using synthetically generated training data using hindsight relabeling.
In contrast to us, the DSL~\cite{hodel2023domain} does not come with a combinatorial search, hence relying for the full search process on the ML model.
Also, since the used transformer is several orders of magnitude larger than ours, test time adaptation is not possible to do, thus relying more on the capabilities of their pre-trained model.

The concurrent work~\cite{akyuerek2024surprising} proposes, similarly to us, to use test-time training for ARC.
They use an 8B-parameter LLM, but due to the great computational demands this work does not allow for 30-minute per task inference.

\subsection{DSL \& Combinatorial}
The ARC challenge has inspired several dedicated DSLs for desribing its transformations, sitting at different levels of abstraction and representational power.

The graph based approach~\cite{xu2023graphs} decomposes a transformation into a series of filters and transformation primitives, that together work on a more abstract graph representation.
It comes together with a combinatorial search algorithm that finds the right overall transformation.
In its basic form it is somewhat limited in its representational power by a small set of elementary transformation primitives, since otherwise the combinatorial search would have difficulties traversing a larger search space.

Ainosoon et al~\cite{ainooson2023approach} propose an imperative style DSL for ARC and tasks are solved through bread-first depth search on a search tree.

The work \cite{lei2024generalized} introduces a novel DSL inspired by \cite{xu2023graphs} and frames the ARC Challenge as a generalized planning problem.
By incorporating graph abstraction~\cite{xu2023graphs}, their solution prunes actions extensively to streamline the search process. Their approach closely resembles ours. However, while we expand the DSL to support a broader search space, their extension remains minimal, likely due to challenges in finding correct solutions as the search space grows. Additionally, they evaluate their solution on only a small subset of ARC tasks.

The work~\cite{ferre2024tackling} provides a DSL written as functional programs in OCaml and searches heuristically for solutions using minimum description length as quality criterion.
It reaches state of the art results for pure DSL-based methods.

Michael Hodel's DSL~\cite{hodel2023domain} decomposes a transformation into a series of 165 elementary operations.
Hand-written programs are given that can solve all of the 400 ARC train tasks.
However, to describe the most complex transformations found in the ARC train set, sequences of up to 60 elementary transformations are necessary.
This can limit the practical effectivity of the DSL since such long programs might be currently too complex for ML models to generate.

\subsection{Simplified ARC Benchmarks}
Due to the complexity of the original ARC benchmark proposed by \cite{chollet2019measureintelligence}, several authors recommend first focusing on simpler tasks, as some of the original ARC tasks are challenging even for humans. The work in \cite{xu2023llms} introduces the 1D-ARC benchmark, which simplifies ARC tasks by reducing them to one dimension, making object association easier to analyze by limiting relationships to right-left interactions. Meanwhile, \cite{moskvichev2023conceptarc} presents ConceptARC, designed to simplify ARC tasks so that nearly all are easily solvable by humans, maintaining two-dimensional images but with an average human success rate of 95\%. Although our paper does not include evaluations on these simplified ARC benchmarks, it underscores the difficulty of the original ARC tasks.

%% file: sec/method.tex
\section{Method}
\label{sec:method}

Our method consists of three parts.
\begin{description}
    \item[DSL:] The domain-specific language (DSL) for encoding image transformations.
    \item[Proposal Generation:] We train a transformer model to predict which specific transformations in the DSL need to be used to get the input/output relationships on the few-shot examples of the current task,
    \item[Combinatorial Search:] for finding the exact subset and order of transformations and their parameters.
\end{description}

Our methodological contribution lies in
(i)~extending ARGA's DSL by additional transformation primitives, enlarging its representational capacity.
(ii)~Introducing a transformer-guided search via proposing transformation primitives.
(iii)~Generating synthetic training examples by hindsight relabeling for pre-training and test-time fine-tuning the transformer.

\subsection{DSL \& Symbolic Engine}
\begin{figure*}[ht]
    \centering
    \includegraphics[width=0.99\textwidth]{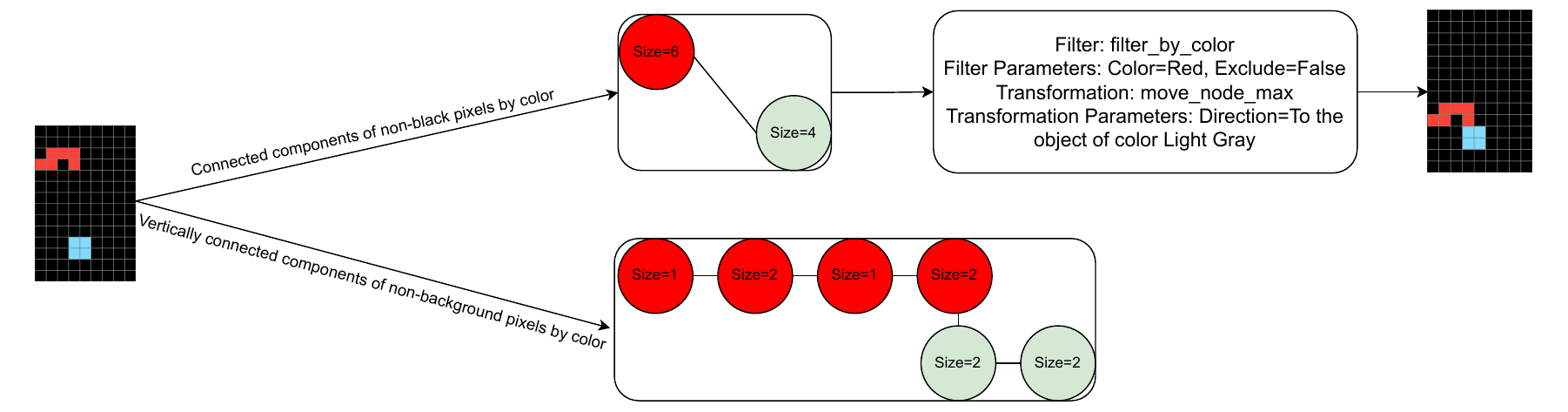}
    \caption{
Two distinct abstractions of the same grid. The upper abstraction groups adjacent pixels of the same color as a single node, while the bottom abstraction groups vertically adjacent components as a single node. Each abstraction also uses a different method for associating nodes within the graph. The image includes the filter, filter parameters, transformation, and transformation parameters needed to achieve the desired output. Note that the choice of abstraction is crucial, since the bottom abstraction will not lead to a correct solution.}
\label{fig: arga_showcase}
\end{figure*}

We use the ``Abstract Reasoning with Graph Abstractions'' (ARGA) DSL from~\cite{xu2023graphs}.
The ARGA DSL consists of three main objects: (i)~abstract representations, (ii)~filters and (iii)~transformation primitives. 
\begin{description}
    \item[Abstract representations] describe the input images in terms of a more abstract graphical structure.
    ARGA associates connected components of the input pixel grids w.r.t.\ different connectivity criteria as graph nodes.
    For example, a node might be a same-color 4-connected subset of the pixel grid.
    Edges are added according to various spatial proximity criteria.
 Note that there can be multiple abstract representations of the same image. In Figure~\ref{fig: arga_showcase} we show two different graph abstractions of the same image.
    \item[Filters] select certain nodes in the abstraction graph.
    Possible filter operations are to select by size, color, neighborhood relations etc.
    \item[Transformation primitives] modify input nodes selected by filters. Exemplary transformation primitives are update color of a node, rotate, extend by appending additional pixels in some directions etc.
\end{description} 

Thus, a full transformation consists of choosing the used abstraction, a sequence of filters and filter parameters and a sequence of transformation primitives and their parameters.
An example of a full transformation is included in Figure ~\ref{fig: arga_showcase}


\begin{figure}[ht]
    \centering
    \includegraphics[width=0.48\textwidth]{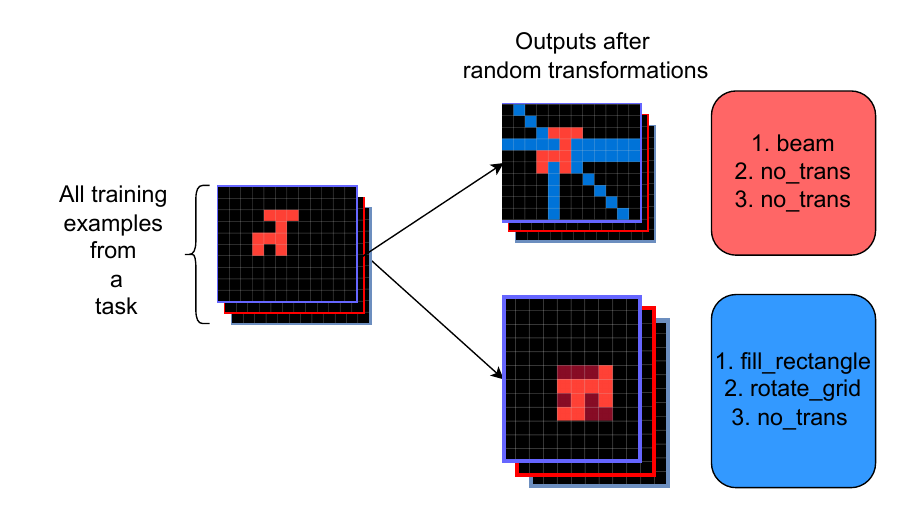}
    \caption{
Hindsight relabeling example: Starting with the original input, we sample abstractions, filters, filter parameters, transformations and transformation parameters. 
We apply them to the grid, generating new input-output pairs with the corresponding transformation.
The labeled pair is added to the transformer's training set.}
\label{fig:hindsight-relabeling}
\end{figure}

ARGA provides 4 base filters out of the box.
More complex filters can be constructed by composing basic ones.
Additionally, 12 transformation primitives are given.

We add 15 more generic and widely usable transformation primitives and call the resulting DSL \emph{extended ARGA}, or abbreviate as \emph{ARGAe}.
The list of additional transformation primitives is given in Table~\ref{tab:additional_transformation_primitives}.

\begin{table}[ht]
\centering
\small 
\setlength{\extrarowheight}{1pt}
\begin{tabular}{m{0.32\columnwidth} m{0.59\columnwidth}} 
\hline
\textbf{Transformation Primitive} & \textbf{Description} \\ \hline
extract & Extract a node \\
duplicate & Duplicate a grid a number of times along an axis \\ 
upscale\_grid & {Multiply grid size by a factor and expand every pixel accordingly} \\
fill & {Fill a subgrid with a specified object} \\
magnet & {Move pixels until they touch a specified border (either image border or a node)} \\
beam & {Shoot colored line in specific directions} \\ 
shift & {Shift the grid in a specified direction} \\
mirror\_duplicate & {Duplicate and mirror the grid} \\ 
rotate\_duplicate & {Duplicate the grid with a specified list of angles} \\
mirror\_grid & {Mirror the grid along a specified axis} \\
rotate\_grid & {Rotate the grid by a specified angle} \\
connect & {Connect two pixels or objects with a line of a specified color} \\
recolor & {Recolor the object to a specified color} \\
truncate & {remove specified object and recolor pixels to background} \\ \hline
\end{tabular}
\caption{Additional transformation primitives in ARGAe.}
\label{tab:additional_transformation_primitives}
\end{table}

\begin{remark}
    Importantly, increasing ARGA's representational capacity by adding more transformation primitives might make it worse in practice.
    Since more transformation primitives mean a larger search space to traverse, some solutions found previously by ARGA's combinatorial search on the smaller base set of transformation primitives might not be found on the enlarged set.
    This arguably is one of the reasons why~\cite{xu2023graphs} did not include a larger set of transformation primitives.
\end{remark}

\paragraph{Combinatorial Search}
ARGA implements a greedy best-first search for traversing a search tree.
Each search tree node corresponds to a set of graphs that were obtained by a partial transformation of the input images.
Search tree nodes are expanded in the order of ascending distance to the output images.
Transformations are pruned by choosing constraints which they are not allowed to violate.
Hashing is used to detect different transformations that result in the same update on the considered images.
A taboo list is used to suppress exploring transformations that have not led to better results.

\subsection{Transformer-Guided Search}
To guide the search process, we train a transformer to generate transformation primitive proposals.
Given several input/output images of a single ARC task, it predicts (i)~how many transformation primitives are used and (ii)~which transformation primitives are used in each step.

\paragraph{Architecture}
We use a standard Transformer-encoder~\cite{vaswani2017attention} with a custom tokenizer.
A token is added for encoding each image color, new image row, new input image and new output image.
We append three classification tokens for predicting a sequence of up to three transformations primitives.
If fewer than three primitives are enough, classification tokens can also predict that no primitive is needed.
For this, the additional \texttt{no\_trans} primitive is used.
We tested different configurations and empirically observed that three primitives are sufficient to solve most ARC tasks that can be represented through ARGAe.

\paragraph{Synthetic ARC Task Generation}
For generating a synthetic task we first randomly sample an ARC task and take only its input images.
Then we randomly sample a sequence of abstractions, filters, filter parameters, transformation primitives and their parameters.
The transformation is applied on all input images from the sampled task.
If on any input image the transformation did not result in an altered output image, we discard the sampled transformation.
After generating synthetic tasks, we also remove duplicate tasks.
See Figure~\ref{fig:hindsight-relabeling} for an illustration of the data generation process.

\paragraph{Pre-Training}
We pre-train the transformer on 31125 synthetically generated tasks with standard categorical cross-entropy.


\paragraph{Test-Time Adaptation}
During inference time, we generate additional synthetic training tasks by sampling random transformations but applying them on the considered input images only.
The pre-trained model is fully fine-tuned.

%% file: sec/experiments.tex
\section{Experiments}
In this section we evaluate the effectiveness of our neuro-symbolic approach on both the train and eval dataset of ARC~\cite{chollet2019measureintelligence}.
We find that our approach can significantly improve upon comparable sota methods on the ARC evaluation set and that both the neural and symbolic parts of our approach are necessary for good performance.

\paragraph{Data Generation}
We pre-train our model on 31125 artificially generated training tasks.
Input images were taken from both the train and eval set, similar to the data generation process from CodeIt~\cite{butt2024codeit}.
We stress that no ground truth transformations were seen during training.
Roughly 40\% of the generated tasks have one transformation primitive, another 40\% have two and 20\% need three.
For tasks using one transformation primitive we balanced the training set so that each transformation primitive was generated similarly often.

\paragraph{Architecture \& Training Details}
Our transformer encoder for the proposal generation has 8 layers, 512 feature dimensions and 2048 feedforward dimensions and 8 attention heads per layer. This gives 25.3 million parameters overall.
We pre-train for 27 epochs with a learning rate of $5e-5$ and batch size 32 using the ADAM optimizer.
For TTA we finetune for 15 epochs with the same parameters.
Our custom tokenizer has a vocabulary size of 45.
During our training we also perform learning rate decay. After every 10 epochs we multiply the learning rate by 0.1.

\paragraph{Our Methods}
We provide three methods corresponding to our individual contributions.
\begin{description}[font=\normalfont]
    \item[\texttt{ARGAe}:] The pure ARGA~\cite{xu2023graphs} DSL approach with extended transformation set. We use the provided combinatorial search algorithm on all possible filters and transformations.
    \item[\texttt{NSA w/o TTA}:] Using transformer-guided search with the pre-trained transformer, we predict up to three consecutive transformation primitives from \texttt{ARGAe}. 
    We have three classification tokens for transformation primitive proposal.
    If the second one predicts no\_trans, meaning only one transformation primitive will be needed, we take the top-5 transformation primitives proposed from the first classification token and feed it to the search algorithm.
    If the third classification token predicts no\_trans, meaning two transformation primitives are needed to describe the overall transformation, we  feed the top-4 transformation primitives of the first two classification tokens to the combinatorial search.
    In the remaining case we feed the top-3 transformation primitive predictions of each classification token to the combinatorial search.
    \item[\texttt{NSA}:] Using transformer-guided grid search with test-time adaptation, for each task, we fine-tune on 2500 training samples generated on the given input for 15 epochs.
    Other parameters are equal to \texttt{NSA w/o TTA}.
\end{description}

\paragraph{Literature Baselines}
We compare against recent SOTA methods that follow a comparable evaluation protocol, that is compute is limited to 30 minutes and evaluation is done on at least some subset of either the ARC train or eval set.
This gives us the original \texttt{ARGA} method~\cite{xu2023graphs} and the other DSL methods~\cite{ferre2021first,ainooson2023approach,ainooson2023approach}.
Among the pure ML-based methods we compare to~\cite{mirchandani2023largelanguagemodelsgeneral}.
Additionally, we compare against the DSL/LLM method \texttt{CodeIt}~\cite{butt2024codeit}.
Hence, we do not compare against methods~\cite{greenblatt2024getting,akyuerek2024surprising,li2024tackling} that do not follow this evaluation protocol or report results on a different task set.

Some comparable methods from the literature have evaluated on subsets of both train and test sets instead of the full ones for various reasons.
For example, \texttt{CodeIt}~\cite{butt2024codeit} uses the train set to choose hyperparameters.
\texttt{ARGA}~\cite{xu2023graphs} only evaluates on promising tasks for which the authors think it has the representational capacity etc.
We also believe that some approaches forego testing on the full train/evaluation set due to the large time investment this would incur.

\paragraph{Evaluation Protocol}
For each task, we allow 30 minutes of computation time.
We report results on each set for comparison.
For TTA, this includes the time to generate new data and finetune the model.
Most other approaches allow for trying three transformations and counting only the potentially correct one.
In contrast to this, but similar to~\cite{mirchandani2023largelanguagemodelsgeneral}, we only make one attempt for each task.

\begin{table}[ht]
\centering
\small 
\resizebox{\columnwidth}{!}{%
\begin{tabular}{lcc}
\toprule
\textbf{Method} & \textbf{ARC Train Set} & \textbf{ARC Eval Set} \\ 
\midrule
\texttt{Ferre} \cite{ferre2021first} & 29/400 & 6/400 \\
\texttt{Ferre} \cite{ferre2024tackling} & \orangec{96/400} & 23/400 \\
\texttt{Ainooson MLE} \cite{ainooson2023approach} & 70/400 & 17/400 \\
\texttt{Ainooson Brute Force} \cite{ainooson2023approach} & \redc{104/400} & 26/400 \\
\texttt{Mirchandani} \cite{mirchandani2023largelanguagemodelsgeneral} & 56/400 & 27/400 \\ 
\texttt{CodeIt} \cite{butt2024codeit} & - & \yellowc{59/400} \\
\texttt{ARGA} \cite{xu2023graphs} & 50/400 & 9/400 \\ 
\midrule
\texttt{ARGAe} & 6/400 & 22/400 \\
\texttt{NSA w/o TTA} & 48/400 & \orangec{63/400} \\
\texttt{NSA} (ours) & \yellowc{78/400} & \redc{75}/400 \\ 
\bottomrule
\end{tabular}%
}
\caption{Results on the training and evaluation datasets of ARC tasks.
Best results are colored red, the second best orange, and the third best yellow.
}
\label{tab: exp}
\end{table}

\paragraph{Results \& Discussion}
In Table \ref{tab: exp} we compare our results with existing baselines from the literature.
We outperform baselines that follow the ARC evaluation protocol by 27\% on the evaluation set and land on the third place on the ARC train set.
Our best-performing method \texttt{NSA} solves 78 tasks on the train set, with 104 being solved by~\cite{ainooson2023approach}.
We solve 75 tasks on the eval set, with 59 being solved by the next-best approach CodeIt~\cite{butt2024codeit}.
Interestingly, we see that also our method \texttt{NSA w/o TTA} outperforms comparable methods on the ARC eval set with 63 tasks solved.
When using TTA in \texttt{NSA} performance is pushed even further.

It is noteworthy, that our approach has similar performance on both train and test set, while most other methods do significantly better on the training set.
Possibly this is due to the nature of the \texttt{ARGAe} DSL, which seems to include transformations that help solve tasks in both evaluate and train.

We can also see the trade-off between more expressive DSL vs.\ more required compute time for the combinatorial search in \texttt{ARGAe} play out differently for train and eval.
\texttt{ARGAe} performance is drastically reduced on the train set as compared to plain ARGA.
It seems that the increased representational capacity has not helped, but rather made the combinatorial search more difficult, thus leading to a lower number of found transformations.
On the eval set, the situation is reversed, with increased representational capacity not hindering the search process as much.
\texttt{NSA} overcomes this trade-off by limiting the search space while retaining \texttt{ARGAe}'s representational capacity.

While TTA gives a significant boost, there is a trade-off between computation spent on the combinatorial search vs.\ more fine-tuning.
In our experince it pays most to invest compute into fine-tuning.
We have observed that performance gains do not plateau after one or two fine-tuning iterations but can continue up to 15 epochs.
Investing more time in data generation and fine-tuning enhances prediction accuracy and increases the likelihood that the transformer identifies the correct transformation primitive. This approach allows our search engine to prioritize finding the correct transformation among the most probable predictions. We find it especially beneficial to prioritize fine-tuning over allocating more time to task-solving.

\begin{figure}[ht]
    \centering
    \includegraphics[width=0.48\textwidth]{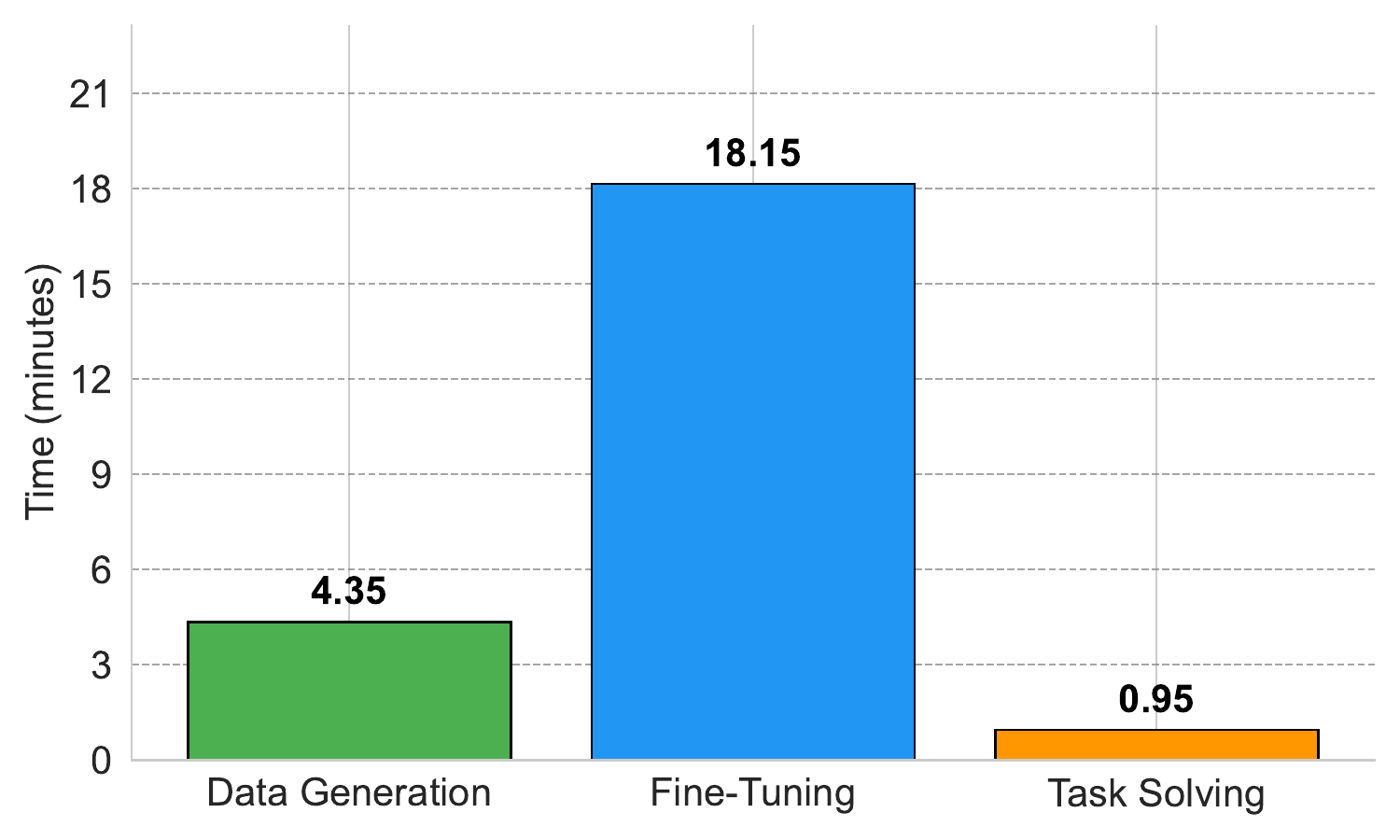}
    \caption{
Comparison of the execution times of \texttt{NSA}'s three components: Data Generation, Fine-Tuning, and Task Solving. We averaged the results across all subsets of tasks we solved. 
}
\label{fig:time}
\end{figure}

Correspondingly, the data-generation and fine-tuning time for each task amounts to around 22 minutes. The average time spent on data generation, fine tuning and task solving for \texttt{NSA} is included in Figure~\ref{fig:time}.

\paragraph{NSA Components Ablations}
In Table \ref{tab: exp} we further examine the necessity of our two key components:
\begin{itemize}
    \item Narrowing the search space by proposal generation through transformer guided search and
    \item Test-time adaptation (TTA) during inference.
\end{itemize}
We see that transformer guided search without TTA, i.e.\ \texttt{NSA w/o TTA} improves upon \texttt{ARGAe} 8-fold on the train set and almost 3-fold on the eval set.
Additionally using TTA in our method \texttt{NSA} further improves by 62\% on the train set and by 20\% on the exal set as compared to no TTA with our method \texttt{NSA w/o TTA}.

\paragraph{TTA Ablations}
We analyze the importance of how long to train and how many data to train on during TTA fine-tuning.
For both ablations we take 50 random tasks overall, such that for 25 of those we can predict the necessary transformation primitives without TTA, while for the rest we cannot do so.

\begin{figure}[ht]
    \centering
    \includegraphics[width=0.48\textwidth]{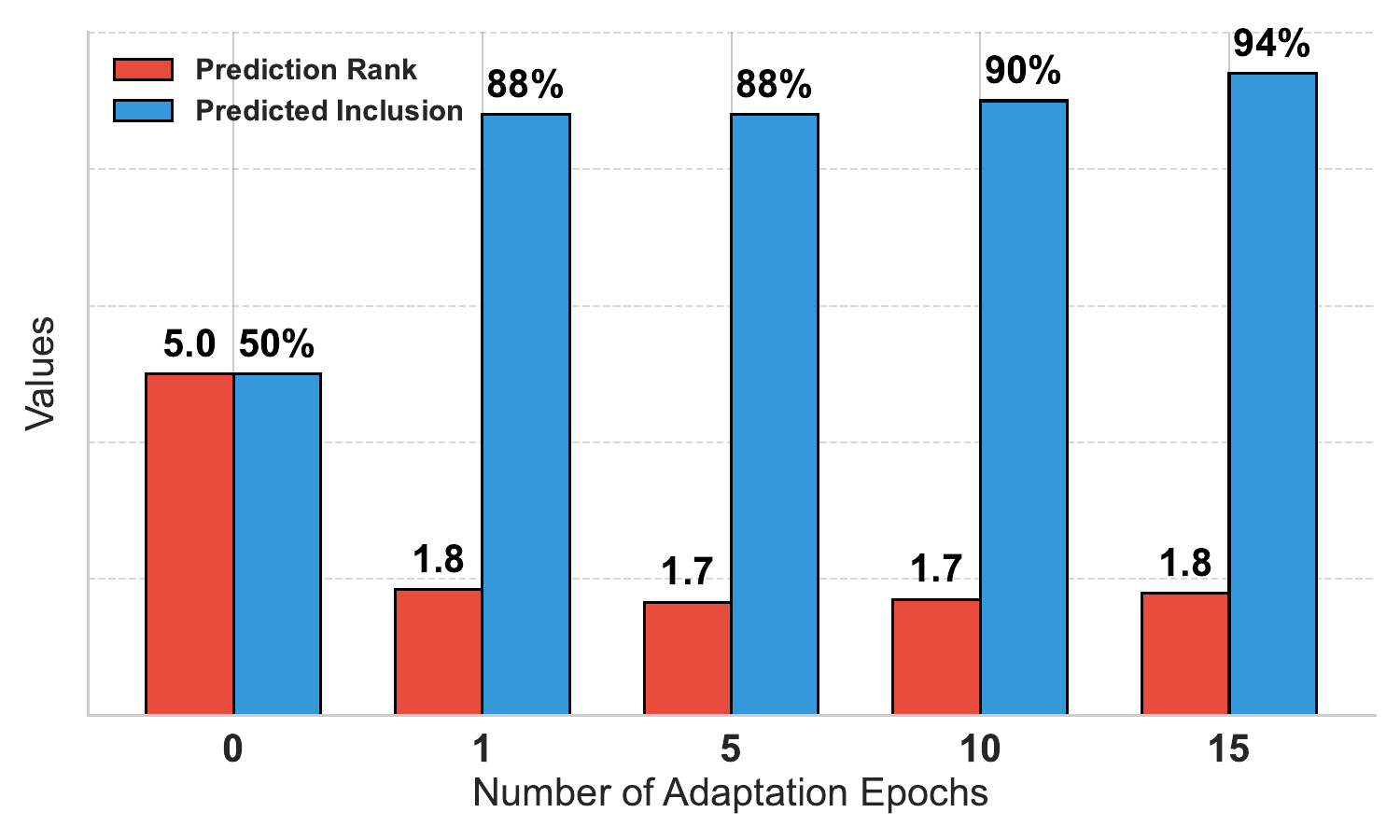}
    \caption{
Prediction Rank and Predicted Inclusion with respect to different TTA Epochs on 50 tasks. Each TTA model was trained using 2500 generated samples.}
\label{fig:albation_tta}
\end{figure}

First, we analyze the influence of the number of TTA epochs on the overall results by evaluating models with 0, 1, 5, 10, and 15 TTA epochs in Figure~\ref{fig:albation_tta}. For this experiment, we provide two metrics:

\begin{itemize} 
\item Prediction Inclusion - This metric indicates whether the true transformation primitives lie within the set of proposed ones. It is essential to assess whether the transformer, after TTA, can correctly predict the transformation primitives at all.
\item Prediction Rank - This metric calculates whether the true transformation primitive has been predicted as most probable (rank 1), second most probable (rank 2) and so on.
Prediction rank is crucial because the search engine prioritizes the most probable transformation primitive first; if the true transformation ranks low, the search may become harder.
\end{itemize}

As shown in Figure \ref{fig:albation_tta}, the average prediction rank does not change much across all TTA epochs.
However, the number of tasks solved fluctuates with different counts of TTA epochs.
Notably, some more difficult tasks yield correct predictions only when using 15 TTA epochs, while easier ones often require just 1 TTA epoch for a correct prediction.
Prioritizing the total number of solved tasks, we choose 15 fine-tuning epochs for our experiments. This figure also highlights the importance of TTA, as even a single epoch significantly boosts the number of tasks with correctly predicted transformation primitives from 50\% to 88\% compared to using no TTA on the ablation tasks.

\begin{figure}[ht]
    \centering
    \includegraphics[width=0.48\textwidth]{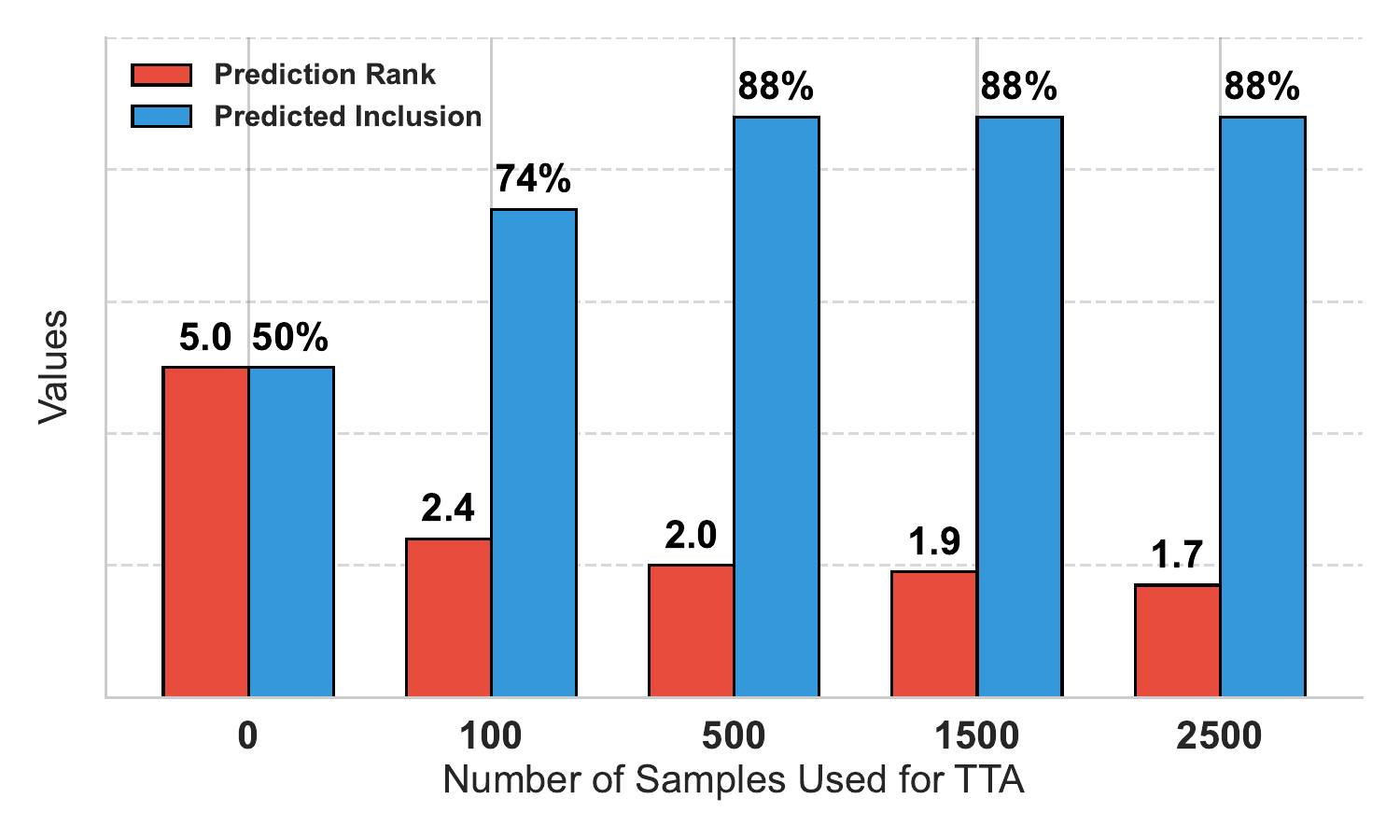}
    \caption{
Prediction Rank and Predicted Inclusion with respect to different number of samples used in TTA for 50 ablation tasks. Each TTA model was estimated for 5 epochs.}
\label{fig:albation_tta_data}
\end{figure}

Second, in Figure \ref{fig:albation_tta_data} we examine the effect of varying the number of TTA training tasks on Prediction Rank and Predicted Inclusion.
We tested sample sizes of 100, 500, 1500, and 2500, along with a baseline comparison of no TTA (denoted as 0 samples in the figure). 
The results show that after 500 epochs Predicted Inclusion stabilizes at 88\%.
Using only 100 already significantly outperforms no TTA, but does not saturate yet.
As the number of TTA samples increases, Prediction Rank improves steadily, suggesting that a larger sample count enables the transformer to make more accurate predictions. 

\label{sec:experiments}

%% file: conclusion.tex
\section{Conclusion}
We have presented a neuro-symbolic approach for ARC challenge.
We show that combining abstraction capabilities of hand-designed DSLs with a learned proposal generation outperforms comparable purely hand-designed search as well as pure ML methods.
One immediate approach is to scale up both the DSL by increasing its representational capacity even more and equally increase the transformer network for the proposal generation, e.g.\ by using pre-trained LLMs and generating larger amounts of training data. However, this approach would soon require more compute than can be provided in 30 minutes.
Another idea is to use several iterations for guessing the correct transformation, iteratively improving the current transformation candidate by inspecting where errors occurred, analoguously to LLM approaches~\cite{yao2022react}.
\section{Acknowledgments}
We extend our gratitude to Adrian Kosmala for his insightful contributions to the project discussions.